\RequirePackage{tikz}
\documentclass[sn-mathphys]{sn-jnl}
\usepackage{bm,libertine}

\theoremstyle{plain}
\newtheorem{theorem}{Theorem}

\newtheorem{example}[theorem]{Example}

\newcommand{\ppull}{\mathsf{pull}}
\newcommand{\ppush}{\mathsf{push}}

\usepackage{color}

\usepackage{tikz}
\usetikzlibrary{arrows.meta}

\usepackage{float}
\usepackage{etoolbox}
\makeatletter
\patchcmd{\ps@headings}
{\hbox to \hsize{\hfill Springer Nature 2021 \LaTeX\ template\hfill}}
{\hbox to \hsize{}}
{}
{}
\patchcmd{\ps@headings}
{\hbox to \hsize{\hfill Springer Nature 2021 \LaTeX\ template\hfill}}
{\hbox to \hsize{}}
{}
{}
\patchcmd{\ps@titlepage}
{\hbox to \hsize{\hfill Springer Nature 2021 \LaTeX\ template\hfill}}
{\hbox to \hsize{}}
{}
{}
\makeatother

\begin{document}

\title{Task Allocation using a Team of Robots}

\author[1]{\fnm{Haris} \sur{Aziz}}\email{haris.aziz@unsw.edu.au}
\author[1,2]{\fnm{Arindam} \sur{Pal}}\email{arindam.pal@data61.csiro.au} 
\author[1]{\fnm{Ali} \sur{Pourmiri}}\email{alipourmiri@gmail.com} 
\author[1]{\fnm{Fahimeh} \sur{Ramezani}}\email{f.ramezani@unsw.edu.au}
\author*[3]{\fnm{Brendan} \sur{Sims}}\email{brendan.sims2@defence.gov.au}
\affil[1]{UNSW Sydney}
\affil[2]{Data61, CSIRO}
\affil[3]{Defence Science and Technology Group}

\abstract{
\textbf{Purpose of Review}: Task allocation using a team or coalition of robots is one of the most important problems in robotics, computer science, operational research, and artificial intelligence. We present a survey of multi-robot task allocation covering  many problem variants and solution approaches, both centralized and decentralized.\\
\textbf{Recent Findings}: In recent work, research has focused on handling complex objectives and feasibility constraints amongst other variations of the multi-robot task allocation problem. There are many examples of important research and recent progress in these directions, which are captured in this survey, along with similar examples for the various solutions that have been developed to solve such problems.\\
\textbf{Summary}: We first present a general formulation of the task allocation problem that generalizes several versions that are well-studied. Our formulation includes the states of robots, tasks, and the surrounding environment in which they operate, and we describe how the problem can be varied depending on the feasibility constraints, objective functions, and the level of dynamically changing information. In addition, we discuss existing solution approaches for the problem including optimization-based approaches, and market-based approaches.
}


\maketitle

\section{Introduction}

Coordinating actions of agents, humans, or robots to allocate and complete tasks is a ubiquitous and fundamental problem. We consider this problem in the context of \emph{multi-robot task allocation (MRTA)}. The general problem has several features. Tasks need to be completed with robots who have the required skills. 
The tasks are heterogeneous, i.e., they are of different types and have different requirements. Depending on the nature of the tasks, the value of completing the tasks may be different. Typically, both the tasks and the robots are in a metric space. Travel costs and energy considerations require that the robots do not undertake tasks that are too far from them. In many of these aspects, the tasks undertaken by robots have an underlying \textit{sequencing} aspect: a robot may be required to undertake tasks in a particular sequence/order. Such decisions are informed by the spatial locations of the robots and tasks. 

In this paper, we survey different variants of the problem and approaches to solve them. Since the problem has been studied across different fields including artificial intelligence, operations research, and combinatorial optimization, several surveys have been written on the topic. We provide an updated survey that has one particular contribution: we present a general model covering several features and constraints related to MRTA. Our formulation helps to encapsulate several MRTA problems that have been extensively studied in the literature. 

\paragraph{Related Surveys and Taxonomies}
Gerkey and Matarić \cite{gerkey2004formal} present a taxonomy for MRTA problems where tasks are assumed to be independent. The taxonomy is structured according to the type of robot, the type of task and the type of assignment. It differentiates between robots that are capable of executing a \textit{single task} at a time ({\bf ST}) and \textit{multiple tasks} simultaneously ({\bf MT}). Tasks are considered according to whether they require a \textit{single robot} ({\bf SR}) or \textit{multiple robots} ({\bf MR}) to complete, and allocations may be \textit{instantaneous} ({\bf IA}) or \textit{time-extended} ({\bf TA}). 
The paper discusses known allocation problems, and some variants that align with the taxonomy along with existing approaches to solve these problems. Korsah et al. \cite{korsah2013comprehensive} extend the taxonomy presented by Gerkey and Matarić \cite{gerkey2004formal} to account for problems involving interrelated utilities and task constraints. Dependencies between tasks are specified as \textit{no dependency} ({\bf ND}), \textit{in-schedule dependencies} ({\bf ID}), \textit{cross-schedule dependencies} ({\bf XD}), and \textit{complex dependencies} ({\bf CD}). These dependencies are applied to Gerkey and Matarić's taxonomy with associated mathematical models and solution approaches discussed.

A further extension to the taxonomy presented by Gerkey and Matarić \cite{gerkey2004formal} is provided by Nunes et al. \cite{nunes2017taxonomy}. They primarily consider temporal and ordering constraints on tasks. They review and organize relevant MRTA literature according to their taxonomy and they discuss both centralized and decentralized solution approaches.

Rizk et al. \cite{rizk2019cooperative} present a survey structured according to the components of the workflow used to automate multi-robot systems (MRS). Coalition formation and task allocation are identified as a key component. Only cooperative coalition formation algorithms are considered, and they are classified according to their assumptions. Various approaches to task allocation in MRS are also discussed. The decision making topology (centralized, decentralized and distributed), heterogeneity of robots and associated applications are highlighted.

Diaz et al. \cite{dias2006market} focus on market-based approaches and review research according to various dimensions. The task allocation problem is considered under the planning dimension. Examples of market-based approaches categorized according to the taxonomy presented by Gerkey and Matarić \cite{gerkey2004formal} are provided along with examples of centralized and distributed mechanisms. There is also a review of approaches that allocate constrained subtasks and approaches that allocate roles.

Yan et al. \cite{yan2013survey} focus on coordination specifically relating to MRS. One key aspect of this coordination is task planning, under which task allocation is considered. A discussion of approaches for MRTA based on the \emph{Contract Net Protocol (CNP)} are presented in addition to various other approaches.

Parker et al. \cite{parker2016multiple} present task allocation as a core concept relevant to the interaction between multiple mobile robot systems and they review behavior-based and market-based task allocation approaches.

The \emph{Multiple Traveling Salesman Problem} is closely related to MRTA and a comprehensive survey is presented by Cheikhrouhou and Khoufi \cite{cheikhrouhou2021comprehensive}. Variants of the problem are articulated and solution approaches are presented in terms of those applied to unmanned ground vehicles and those applied to unmanned air vehicles. The main categories of task allocation approaches considered are deterministic approaches, metaheuristic-based approaches and market-based approaches.

The review by De Ryck et al. \cite{de2020automated} focuses on control algorithms and techniques for automated guided vehicles. The section on task allocation considers relevant task constraints and optimization objectives. The main task allocation approaches reviewed are optimization-based and market-based, whilst there is also a brief mention of behavior-based approaches. 

Khamis et al. \cite{khamis2015multi} provide a comprehensive review on challenging aspects of the MRTA problem. They discuss different MRTA schemes and planning algorithms. They review two popular categories of MRTA solution approaches, namely optimization-based and market-based approaches.
Several other surveys and reviews also include relevant discussion in relation to MRTA \cite{grayson2014search,liu2013robotic,murray2007recent,ota2006multi,skaltsis2021survey}.

\section{Problem Definition}\label{sec:ProblemDefinition}
Let $R=\{r_1,\ldots,r_n\}$ be a set of $n$ \emph{robots} and $T=\{t_1,\ldots,t_m\}$ be a set of $m$ \emph{tasks}. Moreover, for every $r\in R$, define $S_r(\tau)$ to denote the \emph{state} of robot $r$ at time $\tau$. The state of any robot encodes a wide range of information regarding the robot including its position, skills, energy and its current allocation.
Similarly, for every $t\in T$, $S_t(\tau)$ denotes the state of task $t$ at time $\tau$ encoding information such as task constraints and dependencies, robot skill requirements and completion status. Furthermore, $S_e(\tau)$ represents the information regarding a dynamically changing environment including transition costs between tasks and robots and dynamic factors that may affect robot communication. Throughout the paper, we refer to $S_t, S_r, S_e$ for the static case, where the system does not change during the course of an allocation. This case will be the focus of our current discussion. For consideration of dynamic aspects of the problem, please see Section \ref{sec:DynamicSettings}.

In order for a robot $r\in R$ to perform task $t\in T$, it incurs a \emph{cost} $c(S_r(\tau),S_t(\tau),S_e(\tau))$, whilst the \emph{reward} received may be represented as $v(S_r(\tau),S_t(\tau),S_e(\tau))$. Both factors depend on the states. In the case of the cost, this may represent the travel cost to a task, the completion cost of the task and/or some other cost. The combination of cost and reward may be represented in terms of \emph{utility} $u(c,v)$, which seeks to quantify a robot's preference over a particular task. 
The MRTA problem refers to allocating tasks to robots while respecting states and optimizing \emph{objective functions}. The objective functions may be based on any combination of cost, reward, utility or another objective, and are discussed further in Section \ref{sec:Objectives}. 

Both the robots and tasks in MRTA problems may be subject to \emph{constraints}. An in-depth discussion of constraints is included in Section \ref{sec:Constraints}, but here we elaborate on robot capability constraints. A robot may be capable of one or more \emph{skills} $\phi$ and a task may require a robot of a particular skill or multiple robots with a combination of skills. The set of all skills is $\Phi = \{\phi_1,\ldots,\phi_l\}$. Each robot's skill set and the set of skills required by a task are subsets of this set, i.e. $\Phi_r,\Phi_t \subseteq \Phi$.
Using the state-based approach described above, for every robot $r\in R$, a robot's state $S_r$ may be represented as a tuple $(\Phi_r, b_r)$, where $\Phi_r\subseteq \Phi$ is the set of skills the robot is equipped with and $b_r\in \mathbb{R}^+$ denotes the remaining \emph{budget} of robot $r$ to perform tasks. Moreover, for every task $t\in T$, we have $S_t=((\phi, d_{t,\phi}),\ldots)$, where each pair consists of one skill $\phi$ and its corresponding \emph{demand} $d_{t,\phi}$ required by task $t$. In order for robot $r$ to handle required skill $\phi$ of task $t$, the robot needs to have sufficient budget $b_r$ in order to accommodate the demand $d_{t,\phi}$. In what follows we give an example of these particular constraints.

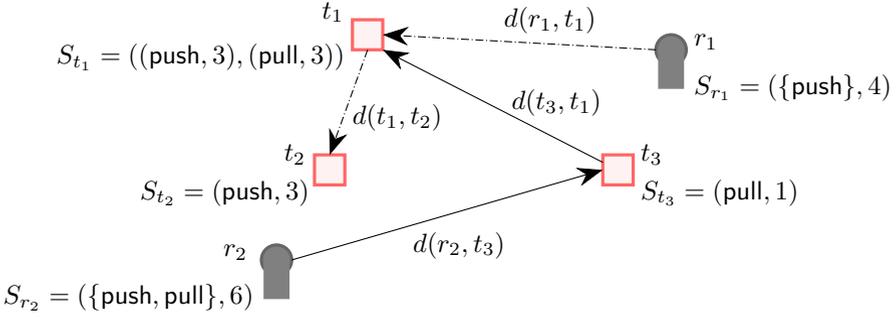
\begin{figure}
\centering
\begin{tikzpicture}
\filldraw[color=black!60, fill=black!50, very thick](-1.2,1.8) circle (0.2);
\filldraw[color=black!50, fill=black!50, very thick](-1.35,1.3) rectangle (-1.05,1.8);
\node[text width=1cm] at (-0.4,1.9) {$r_1$};
\node[text width=5cm] at (1.6,1.3) {$S_{r_1} = (\{\ppush\}, 4)$};
\filldraw[color=black!60, fill=black!50, very thick](-6.4,-1) circle (0.2);
\filldraw[color=black!50, fill=black!50, very thick](-6.55,-1.5) rectangle (-6.25,-1);
\node[text width=1cm] at (-6.6,-0.9) {$r_2$};
\node[text width=5cm] at (-7.5,-1.5) {$S_{r_2} = (\{\ppush, \ppull\}, 6)$};
\filldraw[color=red!60, fill=red!5, very thick](-5.4,1.8) rectangle (-5,2.2);
\node[text width=7cm] at (-5.8,1.7) {$S_{t_1}=((\ppush, 3), (\ppull,3))$};
\node[text width=1cm] at (-5.3,2.3) {$t_1$};
\filldraw[color=red!60, fill=red!5, very thick](-5.9,0) rectangle (-5.5,0.4);
\node[text width=5cm] at (-5.7,-0.1) {$S_{t_2}=(\ppush, 3)$};
\node[text width=1cm] at (-5.8,0.4) {$t_2$};
\filldraw[color=red!60, fill=red!5, very thick](-2.1,0) rectangle (-1.7,0.4);
\node[text width=5cm] at (0.9,-0.1) {$S_{t_3}=(\ppull, 1)$};
\node[text width=1cm] at (-1.1,0.4) {$t_3$};
\draw[-{Stealth[scale=2]}] (-6.2,-1) -- (-2.1,0.2); 
\node[text width =2cm] at (-3.6,-0.8) {$d(r_2,t_3)$}; 
\draw[-{Stealth[scale=2]}] (-2.1,0.3) -- (-5,1.8);
\node[text width =2cm] at (-2.3,1.1) {$d(t_3,t_1)$}; 
\draw[-{Stealth[scale=2]},densely dashdotted] (-1.4,1.8) -- (-5,2);
\node[text width =2cm] at (-2.4,2.2) {$d(r_1,t_1)$};
\draw[-{Stealth[scale=2]},densely dashdotted] (-5.2,1.8) -- (-5.7,0.4);
\node[text width =2cm] at (-4.4,0.9) {$d(t_1,t_2)$}; 
\end{tikzpicture}
\caption{The dashed line indicates a path staring from robot $r_1$ and ending at task $t_2$. The solid line indicates the path starting from robot $r_2$ visiting task $t_3$ and ending at $t_1$. Moreover, $d(r_i,t_j)$ (or $d(t_i, t_j)$) denotes the distance between robot $r_i$ (or task $t_i$) and task $t_j$.}
\label{fig:f1}
\end{figure}

\begin{example}
Suppose that we have a set of two robots, $\{r_1, r_2\}$, and a set of three tasks, $\{t_1,t_2,t_3\}$, scattered on $\mathbb{R}^2$ with standard Euclidean metric denoted by $d(\cdot,\cdot)$ (see Figure \ref{fig:f1}). Let $\Phi=\{\ppush, \ppull\}$ denote the set of skills. Moreover, robot and task states are as follows:
$S_{r_1}=(\{\ppush\}, 4)$, $S_{r_2}=(\{\ppush,\ppull\}, 6)$, 
$S_{t_1}=((\ppush,3),(\ppull,3))$,
$S_{t_2}=(\ppush,3)$, and 
$S_{t_3}=(\ppull,1)$.
For instance, $S_{r_2}$ denotes that robot $r_2$ has skill set $\{\ppush,\ppull\}$ and total budget of six. $S_{t_1}$ indicates that task $t_1$ requires three units of skill $\ppush$ and three units of skill $\ppull$. In order to complete the tasks, each robot visits a subset of tasks and satisfies their demands. Given the skill and budget constraints, task $t_1$ requires both $r_1$ and $r_2$ to be completed. 
As illustrated in Figure \ref{fig:f1}, robot $r_2$ first visits task $t_3$ and then visits $t_1$ whilst robot $r_1$ visits tasks $t_1$ and $t_2$. Notice that each robot incurs a cost which is proportional to travel distance. Therefore, an optimal task allocation assigns an ordered sequence of tasks to each robot and satisfies task requirements while minimizing the total distance travelled by the robots.
\end{example}

\subsection{Constraints on Tasks and Robots}\label{sec:Constraints}

The constraints on robots and tasks in the scope of MRTA restrict how tasks may be allocated to robots.

In order to initially describe task constraints, we assume that tasks are elemental, which refers to singular, non-decomposable tasks. Task constraints may be considered in many dimensions. There may be constraints on the capability of robots completing a task \cite{das2015distributed,emam2020adaptive,mayya2021resilient,Shehory'98}. In Section \ref{sec:ProblemDefinition}, we consider this in terms of skills, i.e., only robots possessing a particular skill are able to complete tasks requiring this skill. There may be requirements on how many robots are required to complete a task \cite{aziz2021multi,gerkey2004formal,michael2008distributed}, i.e., a task may require more than one robot to complete. There may also be dependencies between tasks that dictate that tasks must be completed in a particular order \cite{bischoff2020multi,botelho1999m+,nunes2017taxonomy}, known as precedence or ordering constraints. Furthermore, there may be restrictions on when a task must be completed \cite{choudhury2022dynamic,gini2017multi,nunes2017taxonomy}, known as temporal constraints. For instance, certain tasks may need to be completed simultaneously or a task may need to be completed within a time window. In some cases, there may also be spatial task constraints, where tasks are required to be completed at a certain location \cite{ramchurn2010coalition}. Additionally, there may be task budget constraints, where tasks must be completed within a specified budget \cite{aziz2021multi}, i.e. only robots that can complete the task below a certain cost may be allocated the task.

Instead of elemental tasks, MRTA may consider decomposable tasks, which are represented by a set of subtasks \cite{korsah2013comprehensive}. There may be a set of relationships between these subtasks that dictate the required combination of subtasks for the decomposable task. These relationships may capture any of the constraints described above, or others, that must be satisfied. These decomposable tasks may be allocated as a whole or the subtasks may be allocated separately depending on the nature of the tasks and their constraints.

The constraints on robots may be as varied as those applied to tasks. This includes capability constraints, carrying capacity constraints, budget constraints and spatial constraints amongst others. Capability constraints refer to the case where a robot may only be capable of specific skills, as described above and in Section \ref{sec:ProblemDefinition}. Carrying capacities constrain robots by limiting the number of tasks they may be allocated at a time (e.g., \cite{coltin2014online,ghassemi2022multi}), such as in problems where the tasks involve transportation of items. Alternatively, budget constraints (e.g., \cite{best2018online,luo2015distributed}) may limit the total number of tasks that can be allocated to a robot or the total distance that can be travelled by a robot. Lastly, spatial robot constraints limit where a robot may travel, such as designated start and finish locations \cite{booth2016constraint} or proximity requirements \cite{gombolay2018fast,mosteo2008multi}.

\subsection{Objective Functions}\label{sec:Objectives}

Motivated by a wide range of applications, MRTA seeks to optimize a given set of objective functions, while respecting robot and task constraints. This set of objective functions defines the preferred allocation of tasks to robots. MRTA problems often consider a single objective, but the problem involving multiple objectives represents a more general setting. In this case, the aim is to optimize each unique objective, noting that the preferred allocation may be represented by a \textit{Pareto frontier}.

There are a number of well-studied objective functions considered in the MRTA literature. In certain applications where tasks have associated reward and cost, utility functions may be specified to account for both of these factors \cite{gerkey2004formal,korsah2013comprehensive}. The objective of utility optimization may therefore capture the trade-off between costs and rewards. These objectives may also be considered independently for MRTA problems. If all tasks are equal, cost minimization may be an appropriate objective. This may consider various metrics depending on the application including minimization of distance \cite{Giordani'10, elango2011balancing}, time \cite{mataric2003multi, tang2007complete} or energy \cite{kaleci2013performance,wawerla2010fast}. In the case of time metrics, this could include travel time or task completion time. If tasks have different priority or if robots have preferences over tasks, robots may have distinct values for each task and reward maximization may instead be a suitable objective (e.g., \cite{koes2005heterogeneous}). In some problems, completing all tasks is not feasible. In this case, the objective may relate to completing the maximum number of tasks (e.g., \cite{turner2017distributed}).

Whether the objective relates to utility, cost, reward or something else, there are a number of approaches by which these objectives can be addressed (e.g., see \cite{lagoudakis2005auction}). In the case of cost, we may wish to minimize the total cost of an allocation. Alternatively, we may seek to minimize the maximum cost incurred by any robot, otherwise known as makespan minimization. Finally, we may minimize the average cost of the tasks allocated to robots.

\subsection{Dynamic Settings}\label{sec:DynamicSettings}

Within the context of MRTA, the set of robots, the set of tasks, the operating environment and their associated states may be static and permit an instantaneous allocation or they may be dynamic and change over time requiring a more complex allocation. Here, we acknowledge how dynamic settings may influence the shape of the MRTA problem such that it changes over time. A key factor for MRTA in real-world applications is that the environment may be unknown (e.g., \cite{chaimowicz2002dynamic}), requiring tasks to be re-allocated once more information is revealed. Robots may experience faults that reduce their capabilities (e.g., \cite{parker1998alliance}) or new robots may be introduced (e.g., \cite{ahmadi2006multi}), and tasks may be discovered or arrive over time (e.g., \cite{hoeing2007auction,nunes2015multi}). There may also be uncertainty in a robot's estimation of its ability to complete a task or the requirements of a task (e.g., \cite{emam2020adaptive,emam2021data}). These various dynamic settings may lead to changes in robot preferences over tasks and dictate modification of an allocation or a more dynamic approach to MRTA.

\section{Connections with Well-Studied Problems}
There are a wide range of problems which are closely related to MRTA including job shop scheduling, team orienteering, vehicle routing and coalition formation. We now briefly discuss some of these problems.

\subsection{Vehicle Routing}
The \textsc{Capacitated Vehicle Routing Problem} (CVRP) is one of the most well-studied problems within the class of \textsc{Vehicle Routing Problems} (VRP). In case of CVRP, the transportation requests consist of the distribution of a set of $n$ vehicles, say $R$, from a single \emph{depot} to a set of $m$ customers/tasks, say $T$, which are positioned on a metric space denoted by $S_e$, i.e. the environment state.
The standard definition of CVRP utilizes a weighted graph $G=(V,E,w)$ to represent the environment. The edge weight $w$ indicates the traveling cost/distance from one vertex $V$ to another along edge $E$. In this problem, for every $t \in T$, $S_t$ indicates the capacity demand of each task and, for every $r \in R$, $S_r$ indicates the carrying capacity of vehicle/robot $r$. 
It is assumed that for all $r\in R$, $S_r$ is the same and all vehicles/robots have the same capacity.
 
A vehicle starting from a depot services a subset of customers, and eventually returns to the depot. Therefore, the servicing vehicle $r\in R$ is assigned a closed tour and the cost incurred to vehicle $r$ is defined as
\[
c(r) = \sum \limits_{j=0}^{p-1} c(v_{i_j}, v_{i_{j+1}}) + c(v_{i_p}, v_{i_0}),
\]
where $v_0=v_{i_0}, v_{i_1},\ldots v_{i_p}$ are the vertices being visited by vehicle $r$ in the closed tour.
CVRP seeks to assign a closed tour to each vehicle while minimizing $\sum_{r\in R} c(r)$ and satisfying customer/task demands.

There is an intimate connection between CVRP and MRTA. If we think about the vehicles as robots and the customers visited in going from the source to the destination as tasks, it can be easily seen that CVRP is an instance of MRTA, with the appropriate cost and distance functions.
 
The VRP has a long and fascinating history. Dantzig and Ramser \cite{dantzig1959truck} introduced the VRP as a real-world application concerning the delivery of gasoline to gas stations. They called it the truck dispatching problem. In this seminal paper, they proposed the first mathematical programming formulation and algorithmic approach for the VRP. Clarke and Wright \cite{clarke1964scheduling} proposed an effective greedy heuristic for the approximate solution of the VRP. Gendreau et al. \cite{gendreau1996stochastic} reviewed different types of stochastic vehicle routing problems. Cordeau and Laporte \cite{cordeau2007dial} studied the \textsc{Dial-a-Ride Problem}, which consists of designing vehicle routes and schedules for customers who specify pickup and delivery requests between origins and destinations. The aim is to plan a set of minimum cost vehicle routes capable of accommodating as many customers as possible, under a set of constraints.

\subsection{Coalition Formation}
The coalition formation problem for task allocation is a well-studied setting where each task requires a group of robots of a specified size, i.e. a coalition, to be completed. Moreover, every coalition assigned to a task has a valuation, interpreted as either cost or profit. The goal is to assign a set of coalitions to tasks to optimize a desired objective function with respect to the cost/profit of each coalition. The problem is a variant of the MRTA problem in which, for every task $t\in T$, the task state $S_t$ denotes the number of robots required to perform task $t$. Moreover, the environment state $S_e$ is represented by a complete bipartite graph $G=(R\cup T, E, w)$, where each edge $\{r,t\}$ is associated with a weight $w_{r,t}$. 
$R_t\subset R$ denotes a coalition of robots performing the task $t$. The cost associated with the coalition $R_t$ is $c(R_t) = \sum_{r\in R}w_{r,t}$.
Therefore, the problem seeks to find a collection of coalitions, say $\{R_t: t\in T\}$, which minimizes $\sum_{t\in T} c(R_t)$, and satisfies the demands of all tasks. 

One of the earliest works in this area is by Shehory and Kraus \cite{Shehory'98}. The authors proposed a greedy $k$-approximation algorithm with running time $O(mn^k)$ for $n$ robots, $m$ tasks, and maximum coalition size $k$. Later, the algorithm's running time was improved to $O(mn^{3/2})$ by Service and Adam \cite{Service'11}. Lau et al. \cite{Lau'03} provided a taxonomy and considered different variations of the problem including tasks with unit demands and limited and unlimited resources. Besides the hardness result, polynomial algorithms are presented for special cases.

One can easily see that the problem reduces to the one-to-many bipartite matching problem. Recently, Aziz et al. \cite{aziz2021multi} viewed the model from a budget constraint perspective, where the total task and robot budget is part of the input. The proposed models seek to find an allocation to satisfy the budget constraints.
 
\section{MRTA Solution Approaches}

In this section, we discuss solution approaches for MRTA. In particular, we highlight approaches of note against key elements of our problem definition in Section \ref{sec:ProblemDefinition}. We consider both centralized and decentralized approaches, and we distinguish between the main categories of approaches identified in MRTA literature, namely optimization-based, market-based, and others.
 
\subsection{Optimization-based Approaches}
Optimization-based approaches for MRTA seek to find optimal solutions to problems by searching a solution space. The search is guided by an objective function (or functions) for the problem and the set of solutions is often restricted based on constraints relating to the problem. These approaches typically rely on having access to global information in order to solve the problem. They may be broadly classified as deterministic approaches or stochastic and heuristic approaches.

Although many optimization-based approaches for MRTA are centralised, there are examples of research that demonstrate how these techniques can be applied in a decentralized or distributed manner. Such approaches are particularly important for MRTA in scenarios that lack a central authority to coordinate robot assignments and they avoid the potential single point of failure associated with a central computer. In the following discussion, we include some examples of both centralized and decentralized optimization-based approaches for MRTA.

\subsubsection{Deterministic Approaches}

Deterministic (or exact) approaches for solving MRTA problems are capable of achieving optimal solutions. However, they rely on rigorous algorithms that are typically slow and only suitable for small problem sizes. As the problem size increases, the solution space grows and deterministic algorithms become intractable. In addition, they are less suited for problems with complex constraints. Despite this, a number of deterministic approaches have been developed for MRTA.

A traditional deterministic method for solving allocation problems in polynomial time is the Hungarian method \cite{kuhn1955hungarian}, which is a centralized approach suitable for optimally solving linear assignment problems where robots are allocated single tasks. More recently, distributed versions of the Hungarian method have been proposed for MRTA \cite{chopra2017distributed,Giordani'10}.

Branch-and-bound algorithms, and their variants (e.g. \cite{korsah2012xbots,martin2021multi}), provide improved efficiency over exhaustive search techniques, which consider all possible solutions. These algorithms represent all potential solutions as a tree and search branches of the tree, which correspond to subsets of solutions. Upper and lower bounds on the optimal solution are used to determine which branches are enumerated during the search, i.e. those that can produce better solutions.

When deterministic methods are used to solve MRTA problems, the MRTA problem is often formulated as a mixed integer linear program (MILP) (e.g. \cite{flushing2017simultaneous,koes2005heterogeneous,korsah2012xbots}). In these formulations, the objective functions and constraints are captured with integer and linear equations prior to being solved. Solvers such as CPLEX and Gurobi are often used. MILP formulations for MRTA problems may also be solved with heuristics, which are discussed in Section \ref{sec:StochasticHeuristic}. In addition, MRTA problems may be formulated and solved using constraint programming techniques, where constraints are specified in order to identify feasible solutions, as evidenced by Booth et al. \cite{booth2016constraint}.

\subsubsection{Stochastic and Heuristic Approaches}\label{sec:StochasticHeuristic}

In order to produce solutions for MRTA problems that are much more efficient than deterministic approaches, stochastic and heuristic approaches make use of approximations and heuristics, including metaheuristics. There are many examples of the application of heuristics to solve problems in MRTA scenarios (e.g. \cite{faigl2012goal,mitiche2015efficient,vig2006multi,wawerla2010fast,zhao2015heuristic}). These approximate and heuristic approaches trade-off optimality in order to simplify the search for a solution to optimization problems. Given this, they tend to be more prevalent in MRTA applications.

Metaheuristic approaches guide the search process for solutions to optimization problems. Rather than seeking to enumerate all possible solutions, they only sample a subset of solutions. Unlike more general heuristic approaches, they permit sub-optimal intermediate solutions in order to escape local optima. There are many types of metaheuristic approaches that may be used to solve MRTA problems. They may be classified as trajectory-based and population-based approaches. Other metaheuristics may also be formed that are hybrids of the various types of approaches.

Trajectory-based metaheuristic approaches consider single candidate solutions, which they seek to improve during the search process. Exemplar approaches include simulated annealing (e.g. \cite{mosteo2006simulated,wang2022simulated}) and various search-based approaches, such as Tabu search (e.g. \cite{alighanbari2003coordination,zheng2014tabu}). Simulated annealing involves probabilistically selecting between a current solution and a neighboring solution and is more likely to initially consider worse solutions in order to escape local optima, the probability of which gradually decreases as the solution space is explored. Tabu search uses local search methods to find local optima and considers worse solutions in order to escape these optima, but previously evaluated solutions are avoided through the use of Tabu lists.

Population-based metaheuristic approaches make use of a population of agents to search the solution space and they consider multiple candidate solutions at a time during the search process. There are many types of population-based metaheuristic approaches, including genetic algorithms (e.g. \cite{Choi11,jones2011time,liu2012centralized,martin2021multi,pallin2021decentralized,PatelICRA20}), Particle Swarm Optimization (PSO) (e.g. \cite{manathara2011multiple,wei2020particle}), Ant Colony Optimization (ACO) (e.g. \cite{hu2015hierarchical,wang2012multi,yakici2016solving}), Artificial Bee Colony (ABC) algorithms (e.g. \cite{jevtic2011distributed}) and memetic algorithms (e.g. \cite{chen2018ant,liu2015memetic}) amongst others. Inspired by natural selection, genetic algorithms iteratively evolve candidate solutions through mutation, crossover and selection operations, and a fitness function is used to evaluate solutions until a final solution is obtained. PSO is another iterative search algorithm that moves particles, representing potential solutions, throughout a search space based on position and velocity formulae. A particle's movement is guided by the current best solution found by itself and other particles. ACO adopts pheromone principles used by foraging ants in order to reinforce better solution components as agents iteratively construct candidate solutions from empty solutions. The ABC algorithm is based upon the food seeking behavior of honey bee swarms where food source locations represent potential solutions. Bees (i.e. agents) are guided towards solutions with higher fitness and search neighboring solutions through iterations of the algorithm. Lastly, memetic algorithms seek to combine local search or learning strategies with population-based metaheuristic approaches (e.g. evolutionary algorithms) in order to improve convergence properties.

\subsection{Market-based Approaches}
Market-based solutions to MRTA problems apply market-based principles, such as auctions and negotiation, in order to allocate tasks to robots. One of the earliest works in this space was by Smith \cite{smith1980contract}, where a task-sharing protocol, the so-called \emph{Contract Net Protocol} is proposed. In this protocol, a node, known as the manager, announces an available task that it has generated to the other nodes. Eligible nodes submit their bids for the task to the manager, who identifies and notifies the node that is assigned the task.

The concept of using auctions for assignment problems was first conceived by Bertsekas \cite{bertsekas1979distributed}. Most auction-based algorithms for MRTA apply the principles of CNP to decentralize computation of an allocation of tasks amongst individual robots. Robots express their preferences for tasks with bids, which are shared via explicit communication between robots. The highest or lowest bid on a task typically wins an auction depending on the objective function. In most cases, robots will only have access to local information to formulate their bids, which impairs the ability of an auction to optimize the relevant objective from a global perspective. In an auction, there may be an auctioneer to receive bids and allocate tasks (e.g., \cite{gerkey2002sold,lagoudakis2004simple,simmons2000coordination}) or the winner of an auction may be determined in a distributed manner, such as consensus (e.g., \cite{CBH09a,das2015distributed,luo2011multi}). Winner determination in auctions that make use of an auctioneer is generally simpler, but an auctioneer may represent a single point of failure for the system. However, in some cases, the auctioneer role may be shared between agents (e.g., \cite{lemaire2004distributed,viguria2007set}).

There are many different types of auctions, which are typically differentiated based on the items that are auctioned and the structure of the auction. Tasks may be allocated in single-item auctions (e.g., \cite{koenig2006power,lagoudakis2004simple,lagoudakis2005auction}), multi-item auctions (e.g., \cite{CBH09a,viguria2007set}) or combinatorial auctions (e.g., \cite{berhault2003robot,koenig2007sequential,wen2021multi}). Sequential single-item auctions are relatively simple and only permit the allocation of one task at a time while multiple tasks may be allocated in parallel in a given round in multi-item auctions. Conversely, combinatorial auctions involve combinations of tasks being auctioned as a group. The solutions to combinatorial auctions are more likely to approach the optimal solution, but they require a lot of computation and communication, which grows exponentially as the number of tasks increases. In general, single-item and multi-item auctions are more efficient, but their solutions are less optimal. A comprehensive analysis of various types of auctions for MRTA is provided by Otte et al. \cite{otte2020auctions} who study auction algorithm performance subject to imperfect communication between robots.

Negotiation-based approaches to MRTA work similarly to auctions in that they involve robots evaluating and sharing their preferences for tasks to be allocated (e.g., \cite{botelho1999m+,dias2000free}). However, there is no auctioneer and task allocations are determined independently amongst robots.

In what follows, we seek to highlight and discuss some noteworthy market-based approaches to MRTA.
One of the first market-based approaches developed for MRTA was the M+ protocol \cite{botelho1999m+}. It is a negotiation-based approach where robots incrementally select and negotiate on which tasks to perform after receiving a mission description. However, robots only have local knowledge of the world state. The protocol accounts for task precedence constraints where only executable tasks are able to be selected by robots.

Another early market-based approach developed for MRTA is the MURDOCH system \cite{gerkey2002sold}, which is capable of allocating tasks to robots in an online manner. It uses a simple, greedy auction mechanism, similar to CNP, where one robot acts as an auctioneer for a task introduced to the system. Auctioneers monitor task completion and provide a level of fault tolerance in the event of robot failure.

The approach presented by Zlot et al. \cite{zlot2002multi} proposes a market architecture for multi-robot coordination in exploration scenarios. Robots receive revenue by providing information to a central agent and costs are measured in terms of resources consumed by robots. Robots discover tasks and may opportunistically exchange these tasks with each other via auctions prior to task execution in order to improve efficiency.

Lagoudakis et al. \cite{lagoudakis2005auction} present a theoretical analysis of auction methods for multi-robot routing. The theoretical guarantees offered by various bidding rules and objective functions, specifically approximation ratio bounds, are provided.

The \emph{Consensus-Based Bundle Algorithm (CBBA)} proposed by Choi et al. \cite{CBH09a} is a market-based mechanism that combines auctions and consensus in order to allocate subsets of tasks to robots in a distributed manner. CBBA has become a building block for many decentralized allocation algorithms. The original algorithm relies on submodular bidding functions in order to guarantee convergence, but Johnson et al. \cite{johnson2017decentralized} modified the algorithm in order to achieve improved convergence properties. Motivated by multi-UAV settings, Buckman et al. \cite{buckman2019partial} extended CBBA to a dynamic environment where tasks appear in an online fashion and must be allocated upon arrival. Their proposal enables robots to reset a portion of their previous allocation and replan as tasks arrive. Recently, Chen et al. \cite{chen2022consensus} implemented a replanning approach based on CBBA that considered tasks with timing and capability constraints.

\subsection{Other Approaches}

In this section, we discuss examples of solution approaches not covered in the previous two subsections. This includes learning-based, behavior-based and hybrid approaches amongst others. Hybrid approaches combine different types of approaches in order to solve MRTA problems, which may include optimization-based and market-based approaches. Learning-based approaches incorporate some level of learning to the task allocation process and are often also hybrid solutions. Finally, behavior-based approaches define behaviors for robots and typically involve robots allocating tasks without explicit communication regarding each task.

An example of a learning-based approach for MRTA and scheduling problems is presented by Wang et al. \cite{wang2022heterogeneous}. They propose a heterogeneous graph attention network model in order to learn heuristics for robot scheduling policies in scenarios subject to temporal and spatial constraints. Their approach is applied in the context of final assembly manufacturing with robot teams.

Liu et al. \cite{liu2022option} present a reinforcement learning method for a cooperative multi-robot system. The problem they solve is a multi-task scheduling problem in the context of aircraft painting applications where they seek to avoid collisions between robots. They solved the planning problem with a multi-agent reinforcement learning algorithm based on an option framework, which is an extension of the Markov decision process (MDP). Furthermore, they demonstrate how their proposed solution is capable of handling dynamic situations, such as robot failures.

Park et al. \cite{park2021cooperative} propose a novel MDP formulation for multi-robot task allocation using reinforcement learning in problems involving multi-robot tasks. The problem considered involves sequentially allocating robots to spatial tasks with particular workload requirements and their deep reinforcement learning method makes use of a cross-attention mechanism to compute the importance of tasks for robots. Their solution was compared against various metaheuristics and was shown to outperform these approaches, especially for complex problems.

Schneider et al. \cite{schneider2005learning} present a hybrid approach that combines a market-based approach with learning mechanisms. Tasks are allocated using auctions and learning is applied in order to learn estimated task costs used by robots in bidding for tasks.

Another hybrid approach is presented by Parker and Tang \cite{parker2006building}. They present a behavior-based approach for coalition formation, known as ASyMTRe, which is combined with a negotiation protocol to achieve a distributed version of the algorithm (ASyMTRe-D). ASyMTRe uses schemas, which reside on robots and are combined to achieve particular behaviors in order to complete tasks. ASyMTRe uses a greedy search of schema configurations to identify a solution whilst robots request and obtain information from others via negotiation in ASyMTRe-D.

Other popular behavior-based approaches for MRTA include the ALLIANCE framework developed by Parker \cite{parker1998alliance} and the Broadcast of Local Eligibility (BLE) technique introduced by Werger and Matarić \cite{werger2000broadcast}. In the ALLIANCE framework, robots use motivational behaviors in order to adaptively allocate tasks. Without the use of explicit communication, robots monitor the relative fitness of all robots to perform each task based on their states and the state of the environment. Tasks are represented as behavior sets and the motivation of a robot to select a task will grow over time until it reaches a selection threshold unless it is inhibited by events such as another robot selecting the task. The original framework was also extended in L-ALLIANCE to incorporate learning, which modified the rate of change of motivational behaviors depending on a robot's ability to complete a task. Conversely, BLE involves robots periodically calculating and broadcasting their fitness for completing tasks, which allows each robot to identify and allocate the most eligible robot for completing a particular task.

\section{Conclusion}
In this paper, we systematically surveyed and analyzed the research problems in the field of multi-robot task allocation, focusing on those approaches involving multi-robot coordination. We defined a very general formulation of this problem, encompassing the states of robots, tasks, and the surrounding environment in which they operate. We discussed several constraints on robots and tasks, along with a variety of objective functions. We also considered the problem in terms of dynamic scenarios. Moreover, we studied its connections with well-known problems such as vehicle routing and coalition formation. Finally, we analyzed many solution approaches for MRTA, such as optimization-based, market-based, and some other techniques which do not fall into any of these two categories.

\backmatter

\section*{Acknowledgements}

Aziz, Pourmiri and Ramezani are supported by the Defence Science and Technology Group through the Centre for Advanced Defence Research in Robotics and Autonomous Systems under the project “Task Allocation for Multi-Vehicle Coordination” (UA227119).

\section*{Declarations}

\bmhead{Conflict of Interest} The authors declare that they have no conflict of interest.

\bmhead{Human and Animal Rights and Informed Consent} This article does not contain any studies with human or animal subjects performed by any of the authors.

\renewcommand{\bibpreamble}{Papers of particular interest, published recently, have been highlighted as:\\$\bullet$ Of importance\\}
\bibliography{survey.bib}	
\end{document}